\documentclass[a4paper,12pt]{article}

\usepackage{sbc-template}
\usepackage{graphicx,url}
\usepackage[utf8]{inputenc}

\usepackage{url}   
\usepackage[onelanguage, 
			vlined, 
			linesnumbered, 
			boxed]{algorithm2e} 
\usepackage{amsmath,amsfonts,amssymb}

\newcommand{\xx}{\mathbf{x}}
\newcommand{\XX}{\mathbf{X}}
     
\title{A methodology for detection and localization of fruits \\ 
	in apples orchards from aerial images }

\author{Thiago T. Santos\inst{1}, Luciano Gebler\inst{2} }

\address{Embrapa Informática Agropecuária\\
  Caixa Postal 6041 -- CEP 13083-886 -- Campinas, SP -- Brazil
\nextinstitute
  Embrapa Uva e Vinho\\
  Caixa Postal 177 -- CEP 95200-000 -- Vacaria, RS -- Brazil
  \email{\{thiago.santos,luciano.gebler\}@embrapa.br}
}

\begin{document} 

\maketitle

\begin{abstract}

Computer vision methods based on convolutional neural networks (CNNs) have presented promising results on image-based fruit detection at ground-level for different crops. However, the integration of the detections found in different images,  allowing accurate fruit counting and yield prediction, have received less attention. This work presents  a methodology for automated fruit counting employing aerial-images. It includes algorithms based on multiple view geometry to perform fruits tracking, not just avoiding double counting but also locating the fruits in the 3-D space. Preliminary assessments show correlations above 0.8 between fruit counting and true yield for apples. The annotated dataset employed on CNN training is publicly available. 
 
\end{abstract}
     
\section{Introduction}

Crop monitoring is essential for anomaly detection, yield prediction and 
risk assessment in agriculture, basing the farmer's interventions. A 
continuous data collection during the fruits' growth cycle would allow an 
accurate modeling  of its development, identifying anomalies and bottlenecks. 
Recently, convolutional neural networks \cite{LeCun2015} have been employed
for ground-level, image-based detection for different fruits \cite{Sa2016},
as apples \cite{Hani2020} and mangoes \cite{Bargoti2017}. However, just a few
works \cite{Liu2019,Hani2020,Santos2020} have addressed the \emph{data association} problem 
in fruit counting: how to properly integrate the detections found in multiple images
for accurate, row-level fruit tracking.

The present work describes a methodology for detecting and locating apples 
in orchards from aerial images sequences. This methodology allows not only 
the detection of fruits in the images, but also their association between 
images, identifying apples already observed previously, an essential 
requirement for fruit counting. The identified apples are properly mapped 
in the three-dimensional space, enabling the analysis of the variability in
 the field. The present methodology was able to produce 
 promising results from aerial images of about 1~cm per pixel, 
 thus being an alternative for autonomous monitoring of entire 
 plots in orchards by unmanned aerial vehicles (UAVs).
 
\section{Materials and methods}

The data employed in the development of the methodology came from a plot located at the 
Embrapa's Temperate Climate Fruit Growing Experimental Station 
at Vacaria-RS (28°30'58.2"S, 50°52'52.2"W). The plot, seen in Figure~\ref{fig:sfm}~(a),
is composed of 10 rows of apple trees, of which the 8 inner rows contain the plants of 
 interest (the first and last rows are border ones). The rows contain plants of 
the varieties {\it Fuji} (west facing) and {\it Gala} (east facing). The images were 
taken during December 13, 2018. For aerial shots, an UAV 
(DJI Phantom 4 Pro) performed a 12~m height flight over the orchard's rows, capturing 
imagery data  in the form of a 4K resolution video ($3840 \times 2160$ pixels). 
The camera tilt is not nadir, allowing a more extensive view of the canopy if compared 
to a top/nadir one. The terms \emph{frame} and \emph{image} will be employed interchangeably
in this text.

\subsection{Methodology}

The methodology consists of three steps. The first one is apple detection performed 
on each image, using a deep convolutional neural network \cite{LeCun2015}. The second 
step estimates the camera position and orientation at each frame, using the 
structure-from-motion framework from multiple view computer vision 
\cite{HZ2003, Schoenberger2016}. The last step, the main contribution in this work, 
uses projective geometry and directed graphs to represent multiple alternative 
associations between fruits observed in different frames. Each 
\emph{path} in the graph represents an association hypothesis, determining the location 
of the same fruit in different images, and a greedy algorithm is used to choose the paths.  

\subsubsection{Apple detection}

To this task, we have built an annotated dataset, formed by random selected
$256 \times 256$ pixels samples from the frames extracted from the UAV video 
sequences. The dataset was split in training and test subsets for supervised machine learning, 
as shown in Table~\ref{tab:dataset}. This dataset is publicly available\footnote{Available at \url{https://doi.org/10.5281/zenodo.5586329}.}.

\begin{table}[h]
  \centering
  \caption{Dataset for image-based apple detection training and evaluation.}
  \label{tab:dataset}
  \begin{tabular}{l|rr}
    & Number of images  & Number annotated apples \\ \hline
    Training &  1025 & 2204\\
    Test & 114 & 267 
  \end{tabular}
\end{table}

For apple detection, we have employed a Faster~R-CNN network \cite{RenPAMI2017}, 
using a ResNet-50 backbone \cite{He2016}. We employed the implementation available
in PyTorch \cite{Paszke2019pytorch} (see the \texttt{torchvision} library). The
details of the training process, including data augmentations techniques, optimizer,
batch sizes, number of epochs and hyperparameters can be seen in the publicly
available code\footnote{Available at
  \url{https://github.com/thsant/add256-fastercnn}.} and, due to text
size restrictions, they will not be described here.    

\subsubsection{Relative camera pose estimation}

\begin{figure}
  \centering
  \begin{tabular}{cc}
    \includegraphics[width=0.45\textwidth]{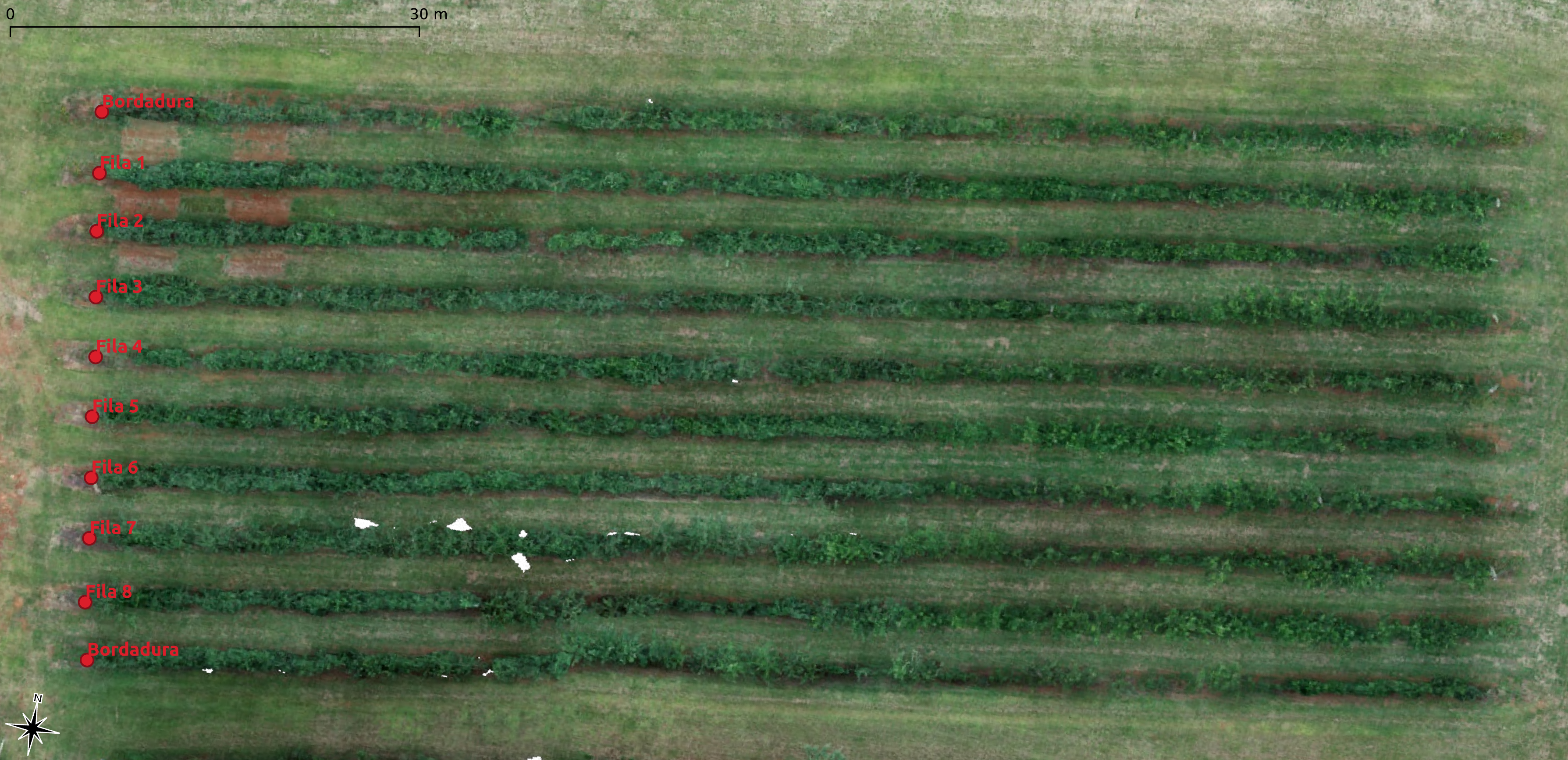} &
    \includegraphics[width=0.45\textwidth]{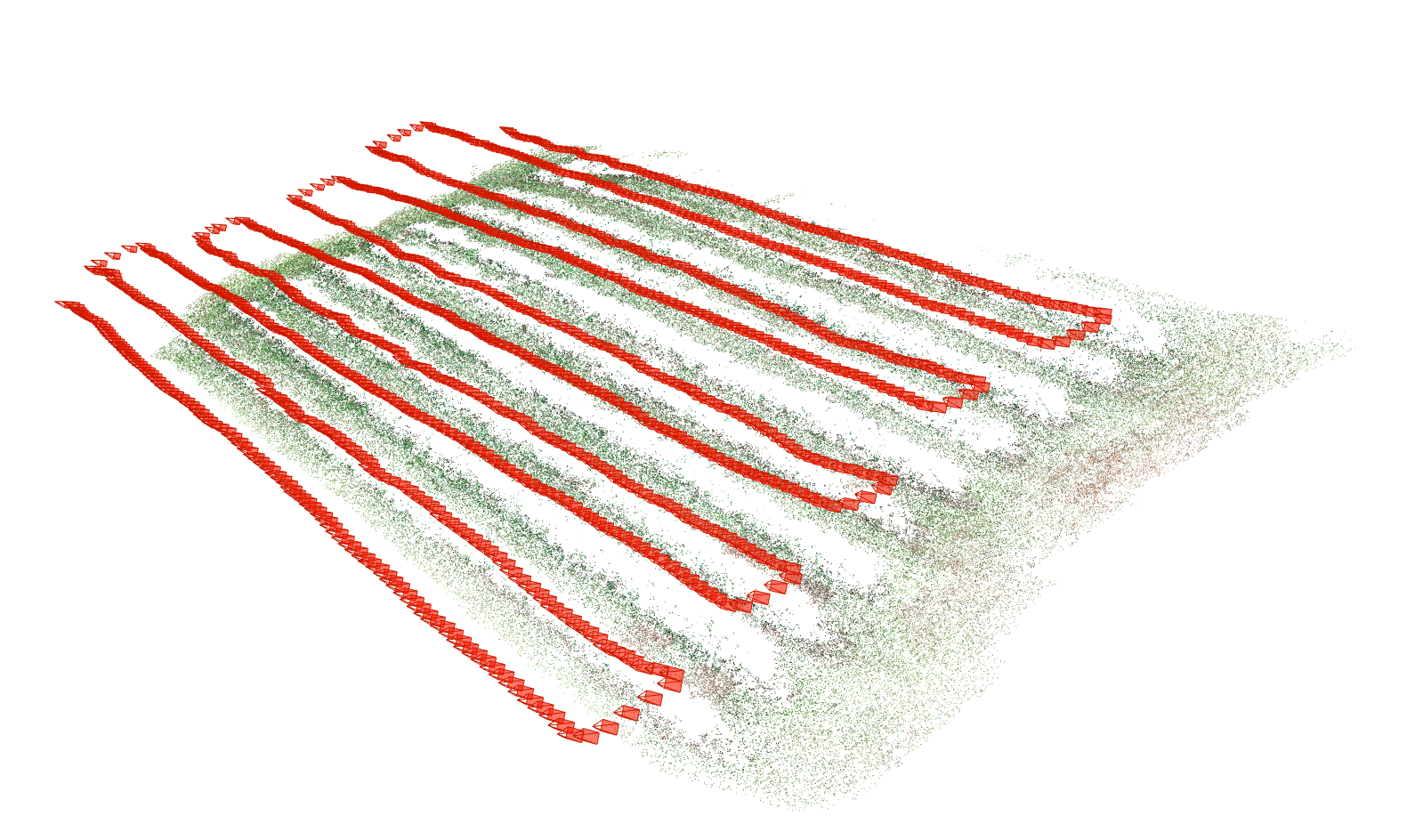} \\
    (a) & (b)
  \end{tabular}
  \caption{The orchard. (a) The plot presenting 10 rows. 
  (b) Structure-from-motion performed by COLMAP - the UAV pose at 
  capture time for each image is shown by the red frustums.}
  \label{fig:sfm}
\end{figure}

To estimate the camera position at the time of capture for each video frame, 
we have employed the Structure-from-Motion (SfM) system COLMAP 
\cite{Schoenberger2016}. A SfM system estimates the \emph{projection matrix} 
$\mathtt{P}_i$, a $3 \times 4$ matrix, for each image $i$: for each  
three-dimensional point $\mathbf{X} = (X, Y, Z, 1)^\intercal$ in the field,
its 2-D projection $\mathbf{x}_i = (x_i, y_i, 1)^\intercal$ on the image
plane of frame $i$ can be computed\footnote{Points $\mathbf{X}$ and $\mathbf{x}_i$ are in
\emph{homogenous coordinates}, what explains the 1 in their last dimension.} by 
the product 
\begin{equation}
  \mathbf{x}_i = \mathtt{P}_i \mathbf{X}.
\end{equation}
The matrices $\mathtt{P}_i$ also allow the computation of the relative position $\mathbf{C}_i$
between cameras in the 3-D space by the property $\mathtt{P}_i \mathbf{C}_i = 0$.
Figure~\ref{fig:sfm}~(b) illustrates the position of 
the UAV camera at the time of each frame capture in the flight over the plot. 


\subsubsection{Data association: tracking apples in the frames sequence}
  
\begin{figure}[h]

  \centering
  \includegraphics[width=\textwidth]{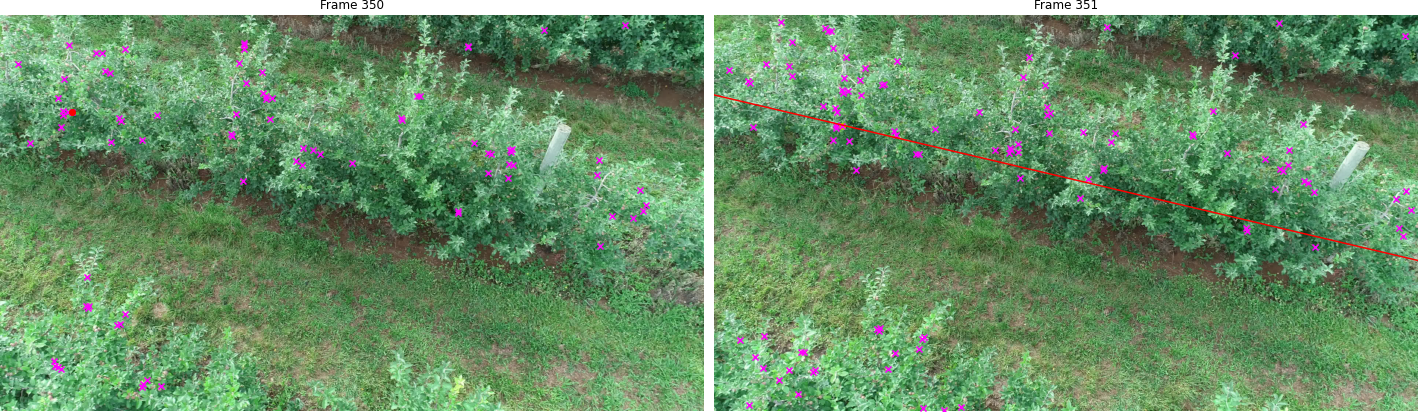}
  \caption{Epipolar restriction. Detected apples are shown as magenta 'x' markers. 
    The point corresponding to the apple marked in red on Frame~350
    defines the red epipolar line seen in Frame~351. The same apple should
    be observed near this line, limiting the number of options for apple
    tracking.} 
  \label{fig:epipolar-constraint}
\end{figure}

Projections matrices $\mathtt{P}_i$ and $\mathtt{P}_j$ allow the computation
of the \emph{fundamental matrix} $\mathtt{F_{i,j}}$ \cite{HZ2003}. Suppose
that a point $\mathbf{X}$ in the 3-D space is mapped to the 2-D points
$\xx_i$ and $\xx_j$ on the $i$-th and the $j$-th frames of the video sequence, 
respectively.
The fundamental matrix maps $\mathbf{x}_i$ on frame $i$ to a
\emph{epipolar line}
$\mathbf{l}_{i,j}$ on frame $j$ that contains $\xx_j$. In our apple tracking problem,
we have $\xx_i^{(m)}$, the centroid of the $m$-th apple detected by 
the neural network on frame $i$. We can employ the fundamental matrix linking 
frames $i$ and $j$ to aid us in choosing the most suitable detections to
correspond to $\xx_i^{(m)}$, as seen in Figure~\ref{fig:epipolar-constraint}. 

Consider the centroids of the $N$ apples detected by the neural network 
on the $j$-th frame, $\xx_{j}^{(n)}, n = 1..N$. The detection corresponding to
apple $\xx_i^{(m)}$ should be close\footnote{Ideally, in a noisy-free, perfect 
detection scenario, $\xx_{j}^{(n)} \in \mathbf{l}_{i,j}^{(m)}$, i.e., 
$\xx_{j}^{(n)} \cdot \mathbf{l}_{i,j}^{(m)} = 0$.}
to the line $\mathbf{l}_{i,j}^{(m)}$ in frame $j$, given by
\begin{equation}
 \mathbf{l}_{i,j}^{(m)} = \mathtt{F_{i,j}} \cdot \xx_i^{(m)}.
\end{equation}
The fundamental matrix can be computed from the projection matrices by
\begin{equation}
  \mathtt{F_{i,j}} = [\mathbf{e}_j]_\times \mathtt{P}_j \mathtt{P}_i^+,
\end{equation}
where $\mathtt{P}_i^+$ is the pseudo-inverse of $\mathtt{P}_i$, and 
$\mathbf{e}_j = \mathtt{P}_j \mathbf{C}_i$ is the \emph{epipole}, with 
$\mathtt{P}_i \mathbf{C}_i = 0$, i.e., $\mathbf{C}_i$ is projection  
center for the camera in frame $i$ \cite{HZ2003}.

Our proposed apple tracking algorithm employs a \emph{graph}, $G$, to
represent multiple fruit associations hypothesis. Each \emph{node} $v_i^{(m)} \in G$ 
corresponds to the centroid $\xx_i^{(m)}$ of the $m$-th apple detected on a frame $i$.
We add an \emph{edge} $v_i^{(m)} \rightarrow v_j^{(n)}$ iif
\begin{equation}
  \mathrm{dist}(\xx_j^{(n)}, \mathbf{l}_{i,j}^{(m)}) = \frac{\xx_j^{(n)} \cdot \mathbf{l}_{i,j}^{(m)}}{\sqrt{a^2 + b^2}} \leq \tau_{\mathrm{epipolar}},   
\end{equation}
being $\mathbf{l}_{i,j}^{(m)} = (a,  b, c)^\intercal =
\mathtt{F_{i,j}} \cdot \xx_i^{(m)}$. In other words, we are testing if
the distance between the point and the epipolar line is below a threshold
$\tau_{\mathrm{epipolar}}$. This
procedure is performed by the lines 4--9 in
Algorithm~\ref{alg:assocdefrutos}, \textsc{FruitAssociation}. So, \emph{an edge
in $G$ represents a possible association between two detections in
different frames}. As seen in line~5, for each frame $i$, the
following $k$ frames are evaluated for associations, what provides
robustness to momentaneous misdetections of a fruit by the neural network.

A sequence of edges $v_i^{(m)} \rightarrow v_j^{(n)} \rightarrow \ldots
\rightarrow v_k^{(o)}$ is a \emph{path}. \emph{Each path represents a possible 
association hypothesis for a fruit detected in frame $i$ and the fruits detected in the 
following frames.} Lines 10-16 in Algorithm~\ref{alg:assocdefrutos} implement a 
path selection process, employing a second algorithm, \textsc{FruitEstimation3D}
(Algorithm~\ref{alg:estimaca}). 

\begin{algorithm}
  \DontPrintSemicolon
  \KwData{The detected apples' centroids $\mathbf{x}_i^{(m)}$ for each frame $i$, $i = 1..F$}
  \KwResult{A set of 3-D points (apples centers) $\mathcal{X} =
    \{\XX_1, \XX_2,\ldots \XX_L\}$ and their tracks}
  \Begin{
    \For{$i \leftarrow 1$ \KwTo $F$}{
      \lForEach{detected apple $\mathbf{x}_i^{(m)}$ in $i$}{Add node $v_i^{(m)}$ to $G$}
    }
    \For{$i \leftarrow 1$ \KwTo $F$}{
      \For{$j \leftarrow i+1$ \KwTo $\min(i+k, F)$}{
        \For{all $\xx_i^{(m)}$ and $\xx_j^{(n)}$}{
          \If{$\mathrm{dist}(\xx_j^{(n)}, \mathbf{l}_{i,j}^{(m)}) \leq \tau_{\mathrm{epipolar}}$}{
            Add the edge $v_i^{(m)} \rightarrow v_j^{(n)}$ to $G$
          }
        }
      }
    }
    $\mathcal{X} \leftarrow \emptyset$\;
  \For{$i \leftarrow 1$ \KwTo $F$}{
    \For{each $v_i^{(m)}$}{
      $\mathbf{X}, \mathcal{I}, r_{\mathcal{I}} \leftarrow$ \textsc{FruitEstimation3D}($G$, $v_i^{(m)}$)\;
      \If{$\mathbf{X} \neq$ NIL}{
        Add $\mathbf{X}$ to $\mathcal{X}$\;
        Remove from $G$ all edges $v_i^{(m)} \rightarrow v_j^{(n)}$ such that $v_i^{(m)}, v_j^{(n)} \in \mathcal{I}$
      }
    }
  }
  \Return{$\mathcal{X}$}
}
\caption{\textsc{FruitAssociation}.}
\label{alg:assocdefrutos}
\end{algorithm}

Algorithm~\ref{alg:estimaca} starts performing a depth-first search (DFS) from
node $v_i^{(m)}$, getting all possible paths starting at $v_i^{(m)}$. 
An algorithm based on \emph{random sample consensus} (RANSAC) \cite{RANSAC} is employed to estimate the tridimensional point $\mathbf{X}$ corresponding to a path (an apple's 3-D position in space). At each
iteration, the \textsc{TriangulationRANSAC} algorithm pick three 2-D points, $\xx_i^{(m)}$, 
$\xx_j^{(n)}$ and $\xx_k^{(o)}$ (corresponding to nodes $v_i^{(m)}$, $v_j^{(n)}$ and 
$v_k^{(o)}$ in a path $T$) and estimates the 3-D point $\mathbf{X}$. The estimation 
of X is performed by a least-squares minimization algorithm
\cite{Hartley1997}. Next, $\mathbf{X}$ is projected on each frame $i$
in the path, defining the points $\hat{\mathbf{x}}_i^{(m)} =
\mathtt{P}_i \mathbf{X}$ and their corresponding  
\emph{geometrical errors}, i.e., the Euclidean distance between $\hat{\mathbf{x}}_i^{(m)}$ and 
$\mathbf{x}_i^{(m)}$. Nodes in the path $T$ whose geometrical error is below the threshold 
$\tau_{\mathrm{geom}}$ are considered \emph{inliers}. At each iteration, the RANSAC 
procedure keeps the point $\mathbf{X}$ that delivered the largest number of inliers.
 Algorithm~\ref{alg:estimaca} looks for the path presenting the largest rate of 
inliers $\mathcal{I}$, keeping the longest path presenting the largest inlier rate. 
In other words, the inlier ratio acts as a quality measure
for the inter-frame association hypothesis regarding fruit $\xx_i^{(m)}$, represented 
by a path starting from $v_i^{(m)}$. Once a path is selected, the algorithm remove its 
edges from $G$ (line 16 in Algorithm~\ref{alg:assocdefrutos}), avoiding those 
associations to be employed again. However, the nodes are preserved in the graph, allowing 
\emph{fruits occlusions} to be considered: fruits that occlude each other can 
create crossing paths in $G$, i.e., paths sharing nodes. 

\begin{algorithm}
  \DontPrintSemicolon
    \KwData{An association graph $G$ and a initial node $v_i^{(m)}$.}
    \KwResult{The 3-D apple position $\mathbf{X}$, a set of inliers 
    nodes $\mathcal{I}$, and the inlier ratio $r_{\mathcal{I}}$.} 
    \Begin{
      $\mathcal{T} \leftarrow$ \textsc{DFS($G$, $v_i^{(m)}$)}\;
      $\mathbf{X} \leftarrow$ NIL\;
      $\mathcal{I} \leftarrow \emptyset$\;
      $r_{\mathcal{I}} \leftarrow 0$\;
      \For{each track $T = \langle v_i^{(m)} \rightarrow v_j^{(n)}
        \rightarrow \ldots \rangle \in \mathcal{T}$, from the longest to the shortest}{
        $\mathbf{X}_T, \, \mathcal{I}_T \leftarrow$ \textsc{TriangulationRANSAC}($T$)\;
        \If{$\mathcal{I}_T \neq \emptyset$}{
          $r_{T} \leftarrow \frac{\Vert\mathcal{I_T}\Vert}{\Vert T \Vert}$\;
          \If{$r_{T} > r_{\mathcal{I}}$}{
            $r_{\mathcal{I}} \leftarrow r_{T}$\;
            $\mathbf{X} \leftarrow \mathbf{X}_T$\;
            $\mathcal{I} \leftarrow \mathcal{I}_T$
          }
        }
      }
      \Return{$\mathbf{X}, \mathcal{I}, r_{\mathcal{I}}$}
    }
    \caption{\textsc{FruitEstimation3D}.}
    \label{alg:estimaca}
\end{algorithm}

\section{Results and discussion}

\begin{figure}[h]
  \centering
  \includegraphics[width=0.75\textwidth]{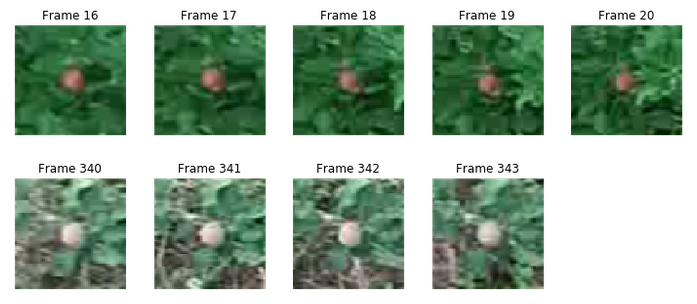}
  \caption{Two inter-frame fruit association examples found by 
  Algorithm~\ref{alg:assocdefrutos}. Each row corresponds to an apple, 
  observed in a few frames. Note the inter-frame variations, caused by UAV's pose 
  changes during recording.} 
  \label{fig:tracking}
\end{figure}

\begin{figure}[h]
  \centering
  \includegraphics[width=\textwidth]{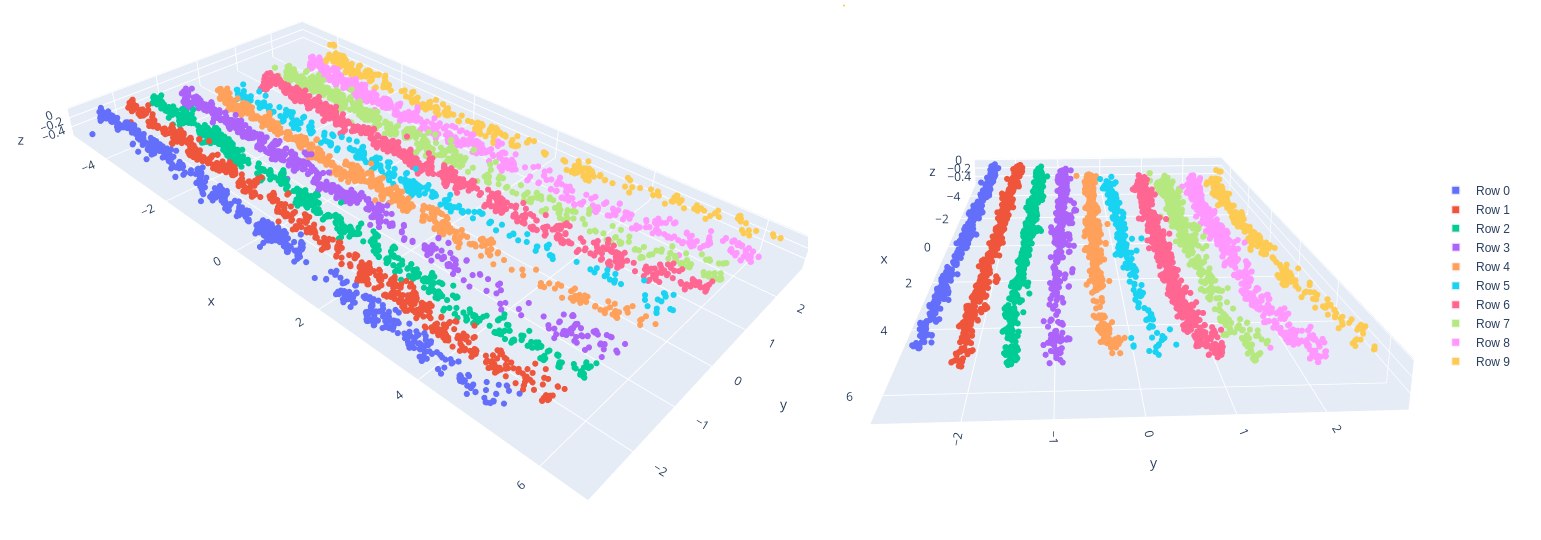}
  \caption{Fruit automatic localization. Each point represents the three-dimensional location determined for an apple. Colors represent different lines in the field, automatically identified using the K-means algorithm.} 
  \label{fig:resultado-vant}
\end{figure}

Figure~\ref{fig:tracking} displays the tracks determined by 
Algorithm~\ref{alg:estimaca} for two different apples. Each line in 
the figure corresponds to an apple's track  
(only the {\it inliers}). Note how the look of the fruit and its
surroundings varies slightly as the pose (the UAV position) changes
from frame to frame. Each track determines the three-dimensional position of 
an apple: all {\it inliers} are used in the final estimation of the fruit's
position $\XX$ in the 3-D space, again by 
employing the least-squares algorithm \cite{Hartley1997}. 
Figure~\ref{fig:resultado-vant} displays a total of 9,237 apples found 
in the plot. Fruits were automatically divided into the ten rows 
of the field by $K$-means clustering.

Disregarding two rows out of the UAV's field of view, caused by imprecision 
in the vehicle positioning system\footnote{Precise flights,
able to keep the plants in the UAV's field of view, can be performed
by vehicles presenting a precise position control, as a Real-Time Kinematic (RTK) Global Navigation Satellite System  (GNSS). Unfortunately, the vehicle used in this work presented an ordinary GNSS system, without the positioning
corrections provided by RTK.},
the observed linear correlation between the counted apples in each row and the row's yield 
was 0.11 for {\it Fuji} and 0.80 for {\it Gala}, considering six rows. However, one
of the rows (row 8) looks like a severe outlier: considering just the other five rows,
linear correlation is 0.93 for {\it Fuji} and 0.88 for {\it Gala}. Although promising, 
the results should be viewed with caution, given that few rows were
evaluated, at a single plot. More extensive experiments are yet necessary for 
a full characterization of uncertainty in yield prediction and the proportion of
the fruits that is visible in imagery. It should also be noted that the images were 
captured in December and the harvest was carried out in February of the following year, which indicates that the methodology has the potential to provide yield estimated in early stages.
Indeed, the presented methodology can be employed as a component of a
more sophisticated yield prediction system.

\section{Conclusions}

Fruit detection and tracking can be, in short term, applied to
yield prediction and crop monitoring. In the long term, precise
detection and 3-D localization can be employed on harvesting
by autonomous agents. Detection and tracking allow autonomous 
agents to estimate their position relative to the fruit, 
so that accurate handling planning can be performed by the machine. 
The three-dimensional localization can also characterize the 
spatial variability of the fruits in the plots, helping on growing
management according to precision agriculture practices. 

The presented methodology is not restricted to aerial images: 
the same algorithms could be adapted to images obtained by ground 
vehicles with embedded cameras. Autonomous aerial vehicles with 
precise positioning control, such as devices 
equipped with RTK GNSS, could be used as a row-scanning system able to
perform automated field monitoring. New experiments, with a greater 
variability of plants, management regimes 
and plant architectures, should be carried out to validate and adapt
the methodology for operation in different scenarios, and provide a
better characterization of the estimation errors in yield prediction. 

\section*{Acknowledgments}

This work was supported by Brazilian Agricultural Research Corporation
(Embrapa) under grant 01.14.09.001.05.04 and by FAPESP under 
grant (2017/19282-7).

\bibliographystyle{sbc}
\bibliography{refs}

\end{document}